\documentclass[11pt,a4paper]{article}
\usepackage{authblk} 
\usepackage[hyperref]{acl2019}
\usepackage{times}

\usepackage{latexsym}
\usepackage{amsfonts}
\usepackage{amsmath}
\usepackage{graphicx,color,xcolor}
\usepackage{bm}
\usepackage{url}
\usepackage{multirow}

\usepackage{xcolor}
\usepackage{booktabs}
\usepackage{subfigure}

\newcommand\blfootnote[1]{%
  \begingroup
  \renewcommand\thefootnote{}\footnote{#1}%
  \addtocounter{footnote}{-1}%
  \endgroup
}

\aclfinalcopy 

\makeatletter
\newcommand{\printfnsymbol}[1]{%
  \textsuperscript{\@fnsymbol{#1}}%
}
\makeatother

\title{Improved Zero-shot Neural Machine Translation \\via Ignoring Spurious Correlations}
\def \nyu{$^\ddag$}
\def \hku{$^\dagger$}
\def \fair{$^\lozenge$}

\author[\fair]{\bf Jiatao Gu*}
\author[\hku]{\bf Yong Wang*}
\author[\fair\nyu]{\bf Kyunghyun Cho}
\author[\hku]{\bf Victor O.K. Li}
\affil[\fair]{Facebook AI Research}
\affil[\hku]{The University of Hong Kong}
\affil[\nyu]{New York University, CIFAR Azrieli Global Scholar}
\affil[\fair]{\tt \{jgu, kyunghyuncho\}@fb.com}
\affil[\hku]{\tt  \{wangyong, vli\}@eee.hku.hk}

\date{}

\begin{document}
\maketitle
\blfootnote{\textbf{*} Equal contribution.}

\begin{abstract}
Zero-shot translation, translating between language pairs on which a Neural Machine Translation (NMT) system has never been trained, is an emergent property when training the system in multilingual settings. 
However, na\"ive training for zero-shot NMT easily fails, and is sensitive to hyper-parameter setting. The performance typically lags far behind the more conventional pivot-based approach which translates twice using a third language as a pivot. In this work, we address the degeneracy problem due to \textit{capturing spurious correlations} by quantitatively analyzing the mutual information between language IDs of the source and decoded sentences. Inspired by this analysis, we propose to use two simple but effective approaches: (1) decoder pre-training; (2) back-translation. These methods show significant improvement ($4\!\sim\!22$ BLEU points) over the vanilla zero-shot translation on three challenging multilingual datasets, and achieve similar or better results than the pivot-based approach.
\end{abstract}

\section{Introduction}

Despite the recent domination of neural network-based models~\cite{sutskever2014sequence,bahdanau2014neural,vaswani2017attention} in the field of machine translation, which have fewer pipelined components and significantly outperform phrase-based systems~\cite{koehn2003statistical}, Neural Machine Translation (NMT) still works poorly when the available number of training examples is limited. 
Research on low-resource languages is drawing increasing attention, and it has been found promising to train a multilingual NMT~\cite{firat2016multi} model for high- and row-resource languages to deal with low-resource translation~\cite{gu2018universal}. 
As an extreme in terms of the number of supervised examples, prior works dug into translation with zero-resource~\cite{firat2016zero,chen2017teacher,lample2017unsupervised,lample2018phrased} where the language pairs in interest do not have any parallel corpora between them.
In particular, \newcite{johnson2016google} observed an emergent property of \textit{zero-shot translation} where a trained multilingual NMT model is able to automatically do translation on unseen language pairs; we refer to this setting as zero-shot NMT from here on.

In this work, we start with a typical degeneracy issue of zero-shot NMT, reported in several
recent works~\cite{arivazhagan2018missing,sestorain2018zero}, that zero-shot NMT is sensitive to training conditions, and 
the translation quality usually lags behind the pivot-based methods which use a shared language as a bridge for translation~\cite{Utiyama:07,cheng2016neural,chen2017teacher}.
We first quantitatively show that this degeneracy issue of zero-shot NMT is a consequence of capturing spurious correlation in the data. 
Then, two approaches are proposed to help the model ignore such correlation: language model pre-training and back-translation. We extensively evaluate the effectiveness of the proposed strategies on four languages from Europarl, five languages from IWSLT and four languages from MultiUN. Our experiments demonstrate that the proposed approaches significantly improve the baseline zero-shot NMT performance and outperforms the pivot-based translation in some language pairs by $2\!\sim\!3$ BLEU points.

\section{Background}
Given a source sentence $x=\{x_1, ..., x_{T'}\}$, a neural machine translation model factorizes the distribution over output sentences $y=\{y_1, ..., y_T\}$ into a product of conditional probabilities: 
\begin{equation}
p(y|x; \theta) = \prod_{t=1}^{T+1} p(y_t| y_{0:t-1}, x_{1:T'}; \theta),
\end{equation}
where special tokens $y_0$ ($\langle \mathrm{bos}\rangle$) and $y_{T+1}$ ($\langle \mathrm{eos}\rangle$) are used to represent the beginning and the end of a target sentence.
The conditional probability is parameterized using a neural network, 
typically, an encoder-decoder architecture based on either RNNs~\citep{sutskever2014sequence,Cho2014a,bahdanau2014neural}, CNNs~\citep{gehring2017convolutional} or the Transformers~\citep{vaswani2017attention}.

\paragraph{Multilingual NMT}
We start with a many-to-many multilingual system similar to \newcite{johnson2016google} which leverages the knowledge from translation between multiple languages. It has an identical model architecture as the single pair translation model, but translates between multiple languages. 
For a different notation, we use $(x^i, y^j)$ where $i, j \in \{0,...,K\}$ to represent a pair of sentences translating from a source language $i$ to a target language $j$. $K+1$ languages are considered in total. 
A multilingual model is usually trained by maximizing the likelihood over training sets $D^{i, j}$ of all available language pairs $\mathcal{S}$. That is:
\begin{equation}
    \max_\theta \frac{1}{|\mathcal{S}|\cdot|D^{i, j}|}\!
    \sum_{\begin{subarray}{c}(x^i, y^j)\in D^{i, j}, (i, j) \in \mathcal{S}\end{subarray}}
    \!\!\mathcal{L}^j_\theta(x^i, y^j),
    \label{eq.best}
\end{equation}
where we denote $\mathcal{L}^j_\theta(x^i, y^j) = \log p(y^j | x^i, \bm{j}; \theta)$.
Specifically, the target language ID $j$ is given to the model so that it knows to which language it translates, and this can be readily implemented by setting the initial token $y_0=j$ for the target sentence to start with.\footnote{
Based on prior works~\cite{arivazhagan2018missing}, both options work similarly. Without loss of generality, we use the target language ID as the initial token $y_0$ of the decoder.
} 
The multilingual NMT model shares a single representation space across multiple languages, which has been found to facilitate translating low-resource language pairs~\cite{firat2016multi,lee2016fully,gu2018universal,gu2018meta}.

\begin{figure*}[t]
    \centering
    \includegraphics[width=\linewidth]{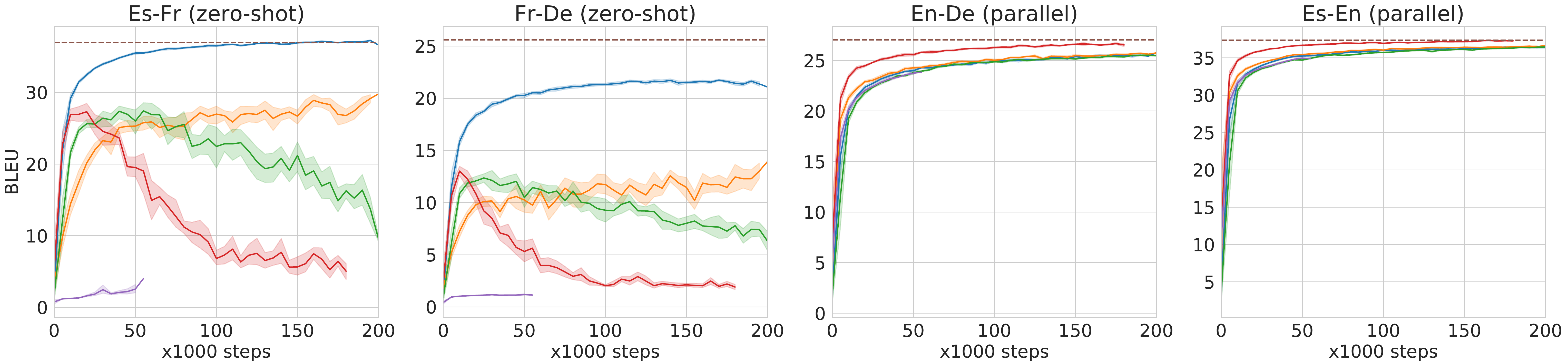}
    \caption{Partial results on zero-shot and parallel directions on Europarl dataset with variant multilingual training conditions ({\color[HTML]{5799c6}blue}: \texttt{default}, {\color{red}red}: \texttt{large-bs}, {\color{orange}orange}: \texttt{pytorch-init}, {\color[HTML]{2da02b}green}: \texttt{attn-drop}, {\color[HTML]{bda1d6}purple}: \texttt{layerwise-attn}). 
    The dashed lines are the pivot-based or direct translation results from baseline models.}
    \label{fig.prelim}
\end{figure*}

\paragraph{Pivot-based NMT}
In practise, it is almost impossible for the training set to contain all $K \times (K+1)$ combinations of translation pairs to learn a multilingual model. Often only one (e.g. English) or a few out of the $K+1$ languages have parallel sentence pairs with the remaining languages. For instance, we may only have parallel pairs between English \& French, and Spanish \& English, but not between French \& Spanish. What happens if we evaluate on an unseen direction e.g. Spanish to French?
A simple but commonly used solution is \textit{pivoting}: we first translate from Spanish to English, and then from English to French with two separately trained single-pair models or a single multilingual model. However, it comes with two  drawbacks: (1) at least $2\times$ higher latency than that of a comparable direct translation model; (2) 
the models used in pivot-based translation are not trained taking into account the new language pair, 
making it difficult, especially for the second model, to cope with errors created by the first model.

\paragraph{Zero-shot NMT}

\newcite{johnson2016google} showed that a trained multilingual NMT system could automatically translate between unseen pairs without any direct supervision, as long as both source and target languages were included in training. In other words, a model trained for instance on English \& French and Spanish \& English is able to directly translate from Spanish to French. Such an emergent property of a multilingual system is called \textit{zero-shot translation}. 
It is conjectured that zero-shot NMT is possible because the optimization encourages different languages to be encoded into a shared space so that the decoder is detached from the source languages. As an evidence,
\newcite{arivazhagan2018missing} measured the ``cosine distance'' between the encoder's pooled outputs of each sentence pair, and found that the distance decreased during the multilingual training.

\section{Degeneracy Issue of Zero-shot NMT}

Despite the nice property of the emergent zero-shot NMT compared to other approaches such as pivot-based methods,
prior works~\cite{johnson2016google,firat2016zero,arivazhagan2018missing}, however, have shown that the quality of zero-shot NMT significantly lags behind pivot-based translation. 
In this section, we investigate an underlying cause behind this particular degeneracy issue.

\subsection{Zero-shot NMT is Sensitive to Training Conditions}    
\label{sec. preliminary experiments}

\paragraph{Preliminary Experiments} 

Before drawing any conclusions, we first experimented with a variety of hyper-parameters to train multilingual systems and evaluated them on zero-shot situations, which refer to language pairs without parallel resource. 

We performed the preliminary experiments on Europarl\footnote{
\url{http://www.statmt.org/europarl/}
} 
with the following languages: English (En), French (Fr), Spanish (Es) and German (De) with no parallel sentences between any two of Fr, Es and De. 
We used newstest2010\footnote{
\url{http://www.statmt.org/wmt18/translation-task.html}
} 
as the validation set which contains all six directions.
The corpus was preprocessed with $40,000$ BPE operations across all the languages. 
We chose Transformer~\cite{vaswani2017attention} -- the state-of-the-art NMT architecture on a variety of languages -- with the parameters of $d_\text{model}=512$, $d_\text{hidden}=2048$, $n_\text{heads}=8$, $n_\text{layers}=6$. Multiple copies of this network were trained on data with all parallel directions for \{De,Es,Fr\} \& En, while we varied other hyper-parameters.
As the baseline, six single-pair models were trained to produce the pivot results. 

\paragraph{Results} 

The partial results are shown in Fig.~\ref{fig.prelim} including five out of many conditions on which we have tested. The \texttt{default} uses the exact Transformer architecture with \textit{xavier\_uniform}~\cite{glorot2010understanding} initialization for all layers, and is trained with $\text{lr}_{\max}=0.005$, $t_\text{warmup}=4000$, $\text{dropout}=0.1$, $n_\text{batch}=2400$ $\text{tokens}/\text{direction}$. For the other variants compared to the \texttt{default} setting, \texttt{large-bs} uses a bigger batch-size of 9,600; \texttt{attn-drop} has an additional dropout (0.1) on each attention head~\cite{vaswani2017attention}; we use the Pytorch's default method\footnote{We use 
\url{https://pytorch.org/docs/master/\_modules/torch/nn/modules/linear.html\#Linear}} 
to initialize all the weights for \texttt{pytorch-init}; 
we also try to change the conventional architecture with a layer-wise attention~\cite{Gu2017NonAutoregressiveNM} between the encoder and decoder, and it is denoted as \texttt{layerwise-attn}. All results are evaluated on the validation set using greedy decoding.

From Fig.~\ref{fig.prelim}, we can observe that the translation quality of zero-shot NMT is highly sensitive to the hyper-parameters (e.g. \texttt{layerwise-attn} completely fails on zero-shot pairs) while almost all the models achieve the same level as the baseline does on parallel directions.
Also, even with the stable setting (\texttt{default}), the translation quality of zero-shot NMT is still far below that of pivot-based translation on some pairs such as Fr-De.

\subsection{Performance Degeneracy is Due to Capturing Spurious Correlation} 
\label{sec. dependency Shifting}

\begin{figure}[!t]
    \centering
    \includegraphics[width=\linewidth]{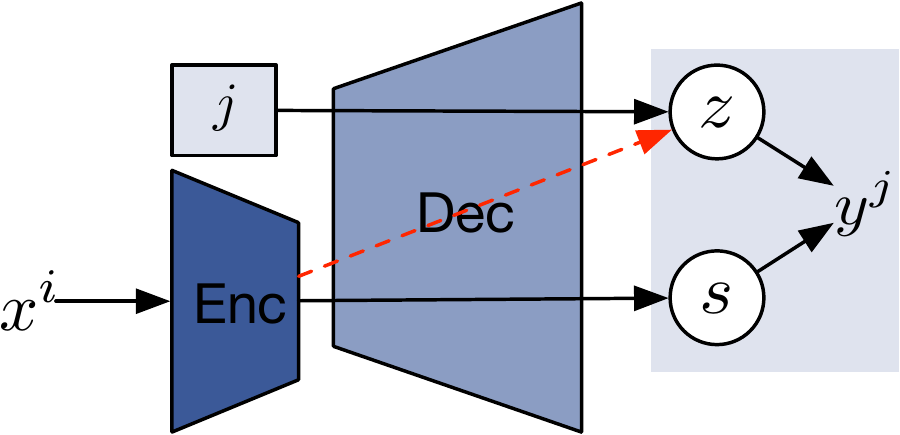}
    \caption{A conceptual illustration of decoupling the output translation ($y^j$) into two latent factors (language type and the semantics) where the undesired spurious correlation (in red) will be wrongly captured if $i$ is always translated to $j$ during training. 
    }
    \label{fig.explain}
 
\end{figure}

We look into this problem with some quantitative analysis by re-thinking the multilingual training in Eq.~\eqref{eq.best}. For convenience, we model the decoder's output $y^j$ as a combination of two factors: the output language ID $z \in \{0,\ldots, K\}$, and language-invariant semantics $s$ (see Fig.~\ref{fig.explain} for a graphical illustration.). In this work, both $z$ and $s$ are unobserved variables before the $y^j$ was generated. Note that $z$ is not necessarily equal to the language id $j$. 

The best practise for zero-shot NMT is to make $z$ and $s$ conditionally independent given the source sentence. That is to say, $z$ is controlled by $j$ and $s$ is controlled by $x^i$. This allows us to change the target language by setting $j$ to a desired language, and is equivalent to ignoring the correlation between $x^i$ and $z$.
That is, the mutual information between the source language ID $i$ and the output language ID $z$ -- $I(i; z)$ -- is \textbf{0}.
However, the conventional multilingual training on an imbalanced dataset makes zero-shot NMT problematic because the MLE objective will try to capture all possible correlations in the data including the spurious dependency between $i$ and $z$. For instance, consider training a multilingual NMT model for Es as input only with En as the target language. Although it is undesirable for the model to capture the dependency between $i$ (Es) and $z$ (En), MLE does not have a mechanism to prevent it (i.e., $I(i; z) > 0$) from happening. 
In other words, we cannot explicitly control  the trade off between $I(i; z)$ and $I(j; z)$ with MLE training. When $I(i; z)$ increases as opposed to $I(j; z)$, the decoder ignores $j$, which makes it impossible for the trained model to perform zero-shot NMT, as the decoder cannot output a translation in a language that was not trained before. 


\paragraph{Quantitative Analysis}

\begin{figure}[t]
    \centering
    \includegraphics[width=\linewidth]{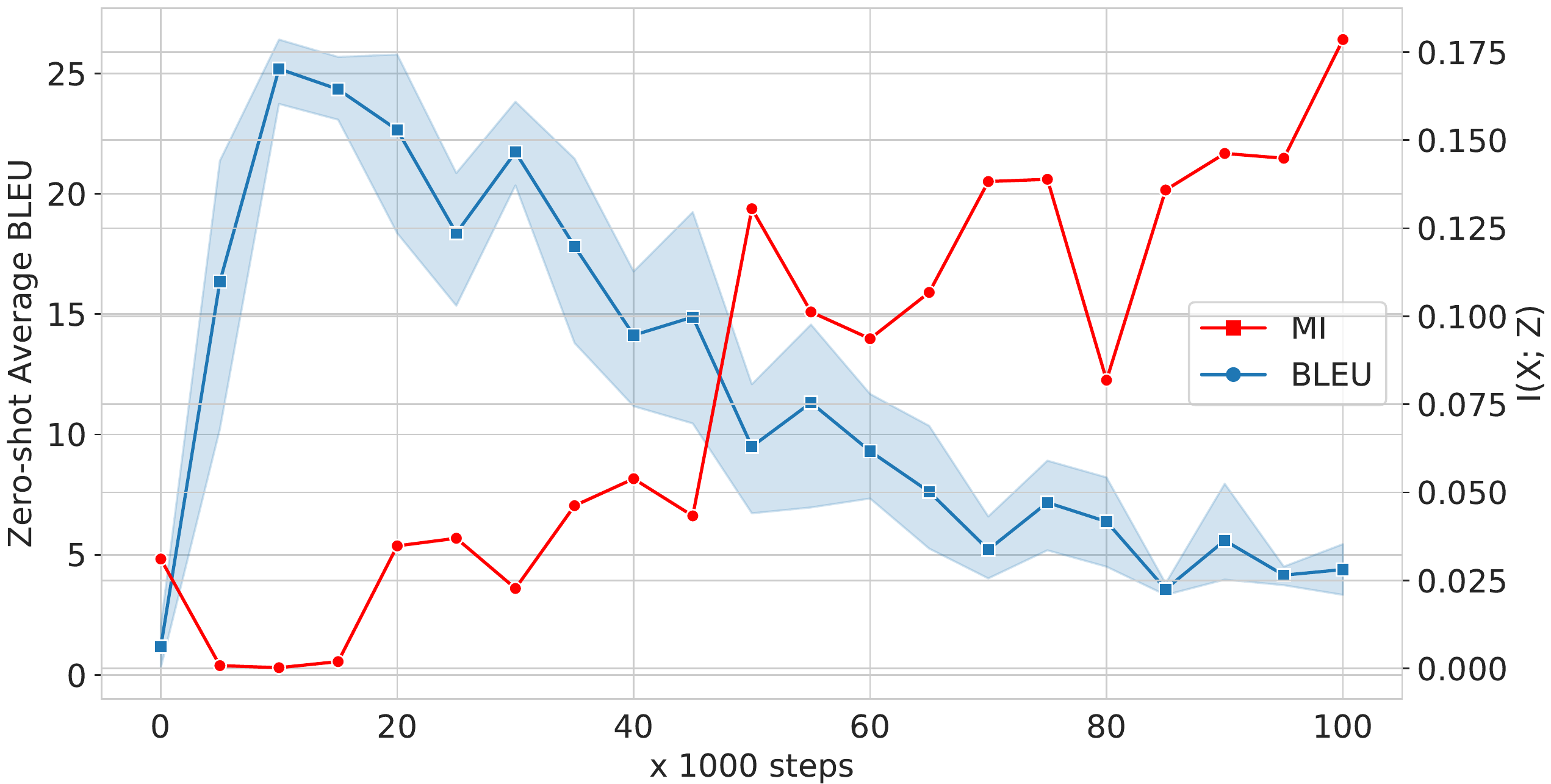}
    \caption{The learning curves of the mutual information between input and output language IDs as well as the averaged BLEU scores of all zero-shot directions on the validation sets for the \texttt{large-bs} setting.}
    \label{fig.mi}
\end{figure}

We performed the quantitative analysis on the estimated mutual information $I(i;z)$ as well as the translation quality of zero-shot translation on the validation set. 
As an example, we show the results of \texttt{large-bs} setting in Fig.~\ref{fig.mi} where the $I(i;z)$ is estimated by:
\begin{equation}
    I(i; z) \approx \frac{1}{(K+1)^2}\sum_{i, j}\log\left[\frac{\tilde{p}(z, i)}{\tilde{p}(z) \cdot \tilde{p}(i)}\right],
\end{equation}
where the summation is over all possible language pairs, and $\tilde{p}(\cdot)$ represents frequency. The latent language identity $z = \phi(y^j)$ is estimated by an external language identification tool
given the actual output~\cite{lui2012langid}. 
In Fig.~\ref{fig.mi},
the trend of zero-shot performance is inversely  proportional to $I(i; z)$, which indicates that the degeneracy is from the spurious correlation.


The analysis of the mutual information also explains the sensitivity issue of zero-shot NMT during training. As a side effect of learning translation, $I(i;z)$ tends to increase more when the training conditions make MT training easier (e.g. large batch-size). The performance of zero-shot NMT becomes more unstable and fails to produce translation in the desired language ($j$).


\section{Approaches}
\label{sec.approach}



In this section, we present two existing, however, not  investigated in the scenario of zero-shot NMT approaches -- decoder pre-training and back-translation -- to address this degeneracy issue.

\subsection{Language Model Pre-training}

The first approach strengthens the decoder language model (LM) prior to MT training. Learning the decoder language model  increases $I(j; z)$ which facilitates zero-shot translation. Once the model captures the correct dependency that guides the model to output the desired language, it is more likely for the model to ignore the spurious correlation during standard NMT training. 
That is, we pre-train the decoder as a multilingual language model. Similar to Eq.~\eqref{eq.best}:
\begin{equation}
    \max_{\theta} \frac{1}{|\mathcal{S}|\cdot|D^{i, j}|}
    \sum_{\begin{subarray}{c}(x^i, y^j)\in D^{i, j}, (i, j) \in \mathcal{S}\end{subarray}}
    \mathcal{\tilde{L}}^j_\theta(y^j),
    \label{eq.best}
\end{equation}
where $\mathcal{\tilde{L}}^j_\theta(y^j) = \log p(y^j | \bm{0}, \bm{j}; \theta)$, which represents that pre-training can be implemented by simply replacing all the source representations by \textbf{zero} vectors during standard NMT training~\cite{sennrich2016edinburgh}.
In Transformer, it is equivalent to ignoring the attention modules between the encoder and decoder. 

The proposed LM pre-training can be seen as a rough approximation of marginalizing all possible source sentences, while empirically we found it worked well. 
After a few gradient descent steps, the pre-trained model continues with MT training. In this work, we only consider using the same parallel data for pre-training.
We summarize the pros and cons as follows:




\vspace{0pt}
\paragraph{Pros:} Efficient (a few LM training steps $+$ NMT training); no additional data needed;
\vspace{0pt}
\paragraph{Cons:} The LM pre-training objective does not necessarily align with the NMT objective.

\subsection{Back-Translation}

In order to apply language model training along with the NMT objective, we have to take the encoder into account. 
We use back-translation~\citep[BT,][]{sennrich2016edinburgh}, 
but in particular for multilingual training. Unlike the original purpose of using BT for semi-supervised learning, we utilize BT to generate \textit{synthetic} parallel sentences for all zero-shot directions~\cite{firat2016zero}, and train the multilingual model from scratch on the merged datasets of both real and synthetic sentences. 
By doing so, every language is forced to translate to all the other languages. Thus, $I(i;z)$ is effectively close to $0$ from the beginning, preventing the model from capturing the spurious correlation between $i$ and $z$.

Generating the synthetic corpus requires at least a reasonable starting point that translates on zero-shot pairs which can be chosen either through a pivot language (denoted as BTTP) or the current zero-shot NMT trained without BT (denoted BTZS). For instance, in previous examples, to generate synthetic pairs for Es-Fr given the training set of En-Fr, BTTP translates every En sentence to Es with a pre-trained En-Es model (used in pivot-based MT), while BTZS uses the pre-trained zero-shot NMT to directly translate all Fr sentences to Es. 
Next, we pair the generated sentences in the reverse direction Es-Fr and merge them to the training set. The same multilingual training is applied after creating synthetic corpus for all translation pairs.
Similar methods have also been explored by \newcite{firat2016zero,zhengmaximum,sestorain2018zero}, but have not been studied or used in the context of zero-shot NMT. 



\vspace{0pt}
\paragraph{Pros:} 
BT explicitly avoids the spurious correlation. Also, BTZS potentially improves further by utilizing the zero-shot NMT model augmented with LM pre-training.
\vspace{0pt}
\paragraph{Cons:} 
BT is computationally more expensive as we need to create synthetic parallel corpora for all language pairs (up to $O(K^2)$) to train a multilingual model for $K$ languages;
both the performance of BTTP and BTZS might be affected by the quality of the pre-trained models.


\begin{table}[t]
	\centering
	\scalebox{0.87}{
	\begin{tabular}{l | l  r  }
	\toprule
	Dataset & parallel pairs & size/pair  \\
	\midrule
    Europarl   & Es-En, De-En, Fr-En  & 2M  \\
    Europarl-b & Es-En, Fr-De & 1.8M \\
    \midrule
    IWSLT   & De-En, It-En, Nl-En, Ro-En & .22M \\
    IWSLT-b & De-En, En-It, It-Ro, Ro-Nl & .22M \\
    \midrule
    MultiUN & Ar-En, Ru-En, Zh-En & 2M \\
    \bottomrule

	\end{tabular}}
	\caption{Overall dataset statistics where each pair has a similar number of examples shown in the rightmost column (we sub-sampled $2M$ sentences per language pair for MultiUN). All the remaining directions are used to evaluate the performance of zero-shot NMT.} 
	\label{table:D1}
	\vspace{0pt}
\end{table}

\section{Experiments}


\subsection{Experimental Settings}

\paragraph{Dataset} 

We extensively evaluate the proposed approaches (LM, BTTP, BTZS) on three multilingual datasets across a variety of languages:
Europarl, IWSLT\footnote{
\url{https://sites.google.com/site/iwsltevaluation2017}
} and MultiUN.\footnote{
\url{http://opus.nlpl.eu/MultiUN.php}
}
The detailed statistics of the training set are in Table~\ref{table:D1},
where we simulate the zero-shot settings by only allowing parallel sentences from/to English. With IWSLT, we also simulate the scenario of having a chain of pivot languages (IWSLT-b).
Also, another additional dataset (Europarl-b) is included where the zero-shot pairs have neither direct nor pivot parallel sentences (similar to unsupervised translation). In such cases, we expect pivot-based methods (including the proposed BTTP) are not applicable.
We use the standard validation and test sets to report the zero-shot performance.
Besides, we preprocess all the datasets following the protocol used in the preliminary experiments. 

\paragraph{Training Conditions} 
For all non-IWSLT experiments, we use the same architecture as the preliminary experiments with the training conditions of \texttt{default}, which is the most stable setting for zero-shot NMT in Sec.~\ref{sec. preliminary experiments}. 
Since IWSLT is much smaller compared to the other two datasets, we find that the same hyper-parameters except with $t_\text{warmup}=8000$, dropout $=0.2$ works better.

\paragraph{Models}
As the baseline, two pivot-based translation are considered: 
\begin{itemize}
    \vspace{0pt}
    \item PIV-S (through two single-pair NMT models trained on each pair;)
    \vspace{0pt}
    \item PIV-M (through a single multilingual NMT model trained on all available directions;)
    \vspace{0pt}
\end{itemize}
Moreover, we directly use the multilingual system that produce PIV-M results for the vanilla zero-shot NMT baseline.

As described in Sec.~\ref{sec.approach}, both the LM pre-training and BT use the same datasets as that in MT training. By default, we take the checkpoint of $20,000$ steps LM pre-training to initialize the NMT model as our preliminary exploration implied that further increasing the pre-training steps would not be helpful for zero-shot NMT. For BTTP, we choose either PIV-S or PIV-M to generate the synthetic corpus based on the average BLEU scores on parallel data. On the other hand, we always select the best zero-shot model with LM pre-training for BTZS by assuming that pre-training consistently improves the translation quality of zero-shot NMT.

\subsection{Model Selection for Zero-shot NMT}
In principle, zero-shot translation assumes we cannot access any parallel resource for the zero-shot pairs during training, including cross-validation for selecting the best model. However, according to Fig.~\ref{fig.prelim}, the performance of zero-shot NMT tends to drop while the parallel directions are still improving which indicates that simply selecting the best model based on the validation set of parallel directions is sub-optimal for zero-shot pairs. 
In this work, we propose to select the best model by maximizing the likelihood over all available validation set $\hat{D}^{i,j}$ of parallel directions together with a language model score from a fully trained language model $\theta'$ (Eq.~\eqref{eq.best}). That is,  
\begin{equation}
    \begin{split}
        \sum_{\begin{subarray}{c}(x^i, y^j)\in \hat{D}^{i, j}\\ (i, j) \in \mathcal{S}\end{subarray}}\!
        \left[\mathcal{L}^j_\theta(x^i, y^j) + \sum_{
        \begin{subarray}{c}(i, k)\notin \mathcal{S} \\ i \neq k\end{subarray}} \frac{\mathcal{\tilde{L}}_{{\theta}'}^k(\hat{y}^k)}{K - |\mathcal{S}|}\right],
    \end{split}
\end{equation}
where $\hat{y}^k$ is the greedy decoding output generated from the current model $p(\cdot|x^i, \bm{k}; \theta)$ by forcing it to translate $x^i$ to language $k$ that has no parallel data with $i$ during training. The first term measures the learning progress of machine translation, and the second term shows the level of degeneracy in zero-shot NMT. Therefore, when the spurious correlation between the input and decoded languages is wrongly captured by the model, the desired language model score will decrease accordingly.

\begin{figure}[t]
    \centering
    \includegraphics[width=0.98\linewidth]{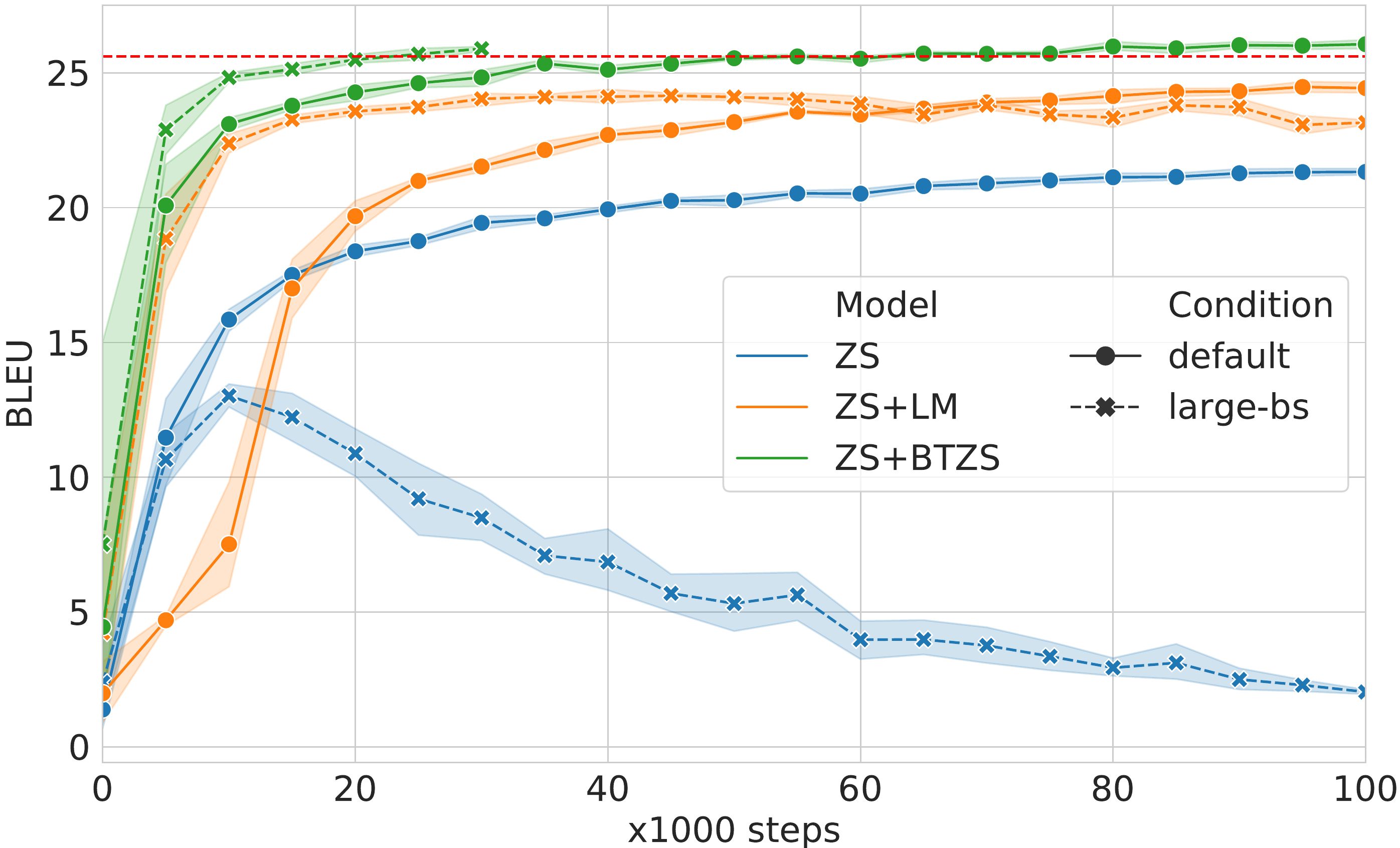}
    \caption{Learning curves of the two proposed approaches (LM, BTZS) and the vanilla ZS on Europarl Fr $\!\rightarrow\!$ De with two conditions (\texttt{default}, \texttt{large-bs}). The {\color{red}red} dashed line is the pivot-based baseline.}
    \label{fig.robustness}
    \vspace{0pt}
\end{figure}

\begin{table*}[t]
	\centering
    \footnotesize
	\subtable{
	\begin{tabular*}{0.99\textwidth}{@{\extracolsep{\fill}} lcc cc cc cc| cc cc}
	\toprule
	Europarl & \multicolumn{8}{c|}{(a) De, Es, Fr $\leftrightarrow$ En} & \multicolumn{4}{c}{(b) Es $\leftrightarrow$ En, Fr $\leftrightarrow$ De}\\
	\midrule
	\multirow{2}{*}{Model} & \multicolumn{2}{c}{De-Es} & \multicolumn{2}{c}{De-Fr} & \multicolumn{2}{c}{Es-Fr} & Zero & Parallel & \multicolumn{2}{c}{Es-Fr} & \multicolumn{2}{c}{De-En} \\
	& $\leftarrow$ & $\rightarrow$ & $\leftarrow$ & $\rightarrow$ & $\leftarrow$ & $\rightarrow$ & Avg & Avg &  $\leftarrow$ & $\rightarrow$ &  $\leftarrow$ & $\rightarrow$\\
	\midrule
    PIV-S            & 26.2 & 31.2 & \textbf{25.9} & 32.2 & 35.7 & 38.0 & 31.5 & \textbf{35.0} &   \multicolumn{4}{c}{ -- not applicable -- }  \\
    PIV-M            & 26.2 & 31.1 & 25.2 & 31.5 & 35.4 & 37.1 & 31.1 & 34.4 &  \multicolumn{4}{c}{ -- not applicable -- } \\
    \midrule 
    ZS                 & 22.1 & 30.2 & 21.7 & 29.6 & 36.2 & 36.7 & 29.4 & 34.4 &   29.5 & 27.5 & 14.3 & 23.7 \\
    ZS+Align~{\shortcite{arivazhagan2018missing}}  & 24.7 & 31.4 & 23.8 & 31.0 & 37.3 & 38.5 & 31.1 & 34.5 &--&--&--&-- \\
    ZS+LM        & 25.9 & 32.8 & 25.5 & 32.3 & \textbf{39.3} & \textbf{40.0} & 32.6 & 34.6 & \textbf{34.9} & \textbf{37.1} & \textbf{21.5} & \textbf{30.0} \\
    ZS+BTTP      & \textbf{27.1} & \textbf{33.0} & \textbf{26.4} & \textbf{33.0} & 39.1 & \textbf{40.0} & \textbf{33.1} & 33.9 & \multicolumn{4}{c}{ -- not applicable -- } \\
    ZS+BTZS      & \textbf{26.7} & \textbf{33.2} & \textbf{25.9} & \textbf{33.1} & \textbf{40.0} & \textbf{41.4} & \textbf{33.4} & \textbf{34.7} & \textbf{39.7} & \textbf{40.5} & \textbf{25.1} & \textbf{30.6} \\
    \midrule
    Full  & 28.5 & 34.1 & 27.9 & 34.2 & 40.0 & 42.0 & 34.4 & 34.8 &  40.0 & 42.0 & 27.0 & 33.4 \\
    \bottomrule
	\end{tabular*}
	}
	
	\subtable{
    \begin{tabular*}{0.99\textwidth}{@{\extracolsep{\fill}} l cc cc cc cc cc cc cc }
	\toprule
	IWSLT & \multicolumn{13}{c}{(c) De, It, Nl, Ro $\leftrightarrow$ En}\\
	\midrule
	\multirow{2}{*}{Model} & \multicolumn{2}{c}{De-It} & \multicolumn{2}{c}{De-Nl} & \multicolumn{2}{c}{De-Ro} & \multicolumn{2}{c}{It-Nl} & \multicolumn{2}{c}{It-Ro} & \multicolumn{2}{c}{Nl-Ro} & Zero  & Parallel   \\
		& $\leftarrow$ & $\rightarrow$ & $\leftarrow$ & $\rightarrow$ & $\leftarrow$ & $\rightarrow$ &  $\leftarrow$ & $\rightarrow$ &   $\leftarrow$ & $\rightarrow$ &  $\leftarrow$ & $\rightarrow$ &  Avg & Avg  \\
	\midrule 
    PIV-S & 16.7 & 16.3 & 19.1 & 17.7 & 17.5 & 15.0 & 18.4 & 18.6 & 18.8 & 17.2 & 18.3 & 17.0 & 17.6 & 29.8   \\
    PIV-M & 21.4 & 21.6 & 24.0 & 23.7 & 22.3 & {20.0} & {22.7} & {22.4} &  23.6 & 21.3 & {23.0} & 21.1 & {22.3} & {35.0} \\
    \midrule 
    ZS       & 14.8 & 17.2 & 16.7 & 17.8 & 14.9 & 16.6 & 18.4 & 16.1 & 19.7 & 17.8 & 16.2 & 17.5 & 17.0 & {35.0}  \\
    ZS+LM  & 21.3 & 20.9 & 24.7 & 24.1 & 22.3 & 19.8 & 22.2 & 22.3 & 23.2 & {22.1} & {23.0} & {21.6} & {22.3} & 34.9   \\
    ZS+BTTP & \textbf{23.3} & \textbf{23.3} & \textbf{26.5} & \textbf{25.8} & \textbf{23.9} & \textbf{22.1} & \textbf{24.6} & \textbf{24.3} & \textbf{25.9} & \textbf{23.7} & \textbf{24.7} & \textbf{23.7} & \textbf{24.3} & \textbf{35.2}  \\
    ZS+BTZS & \textbf{22.6} & \textbf{23.3} & \textbf{27.2} & \textbf{26.5} & \textbf{23.6} & \textbf{21.8} & \textbf{24.3} & \textbf{24.0} & \textbf{25.7} & \textbf{23.6} & \textbf{25.4} & \textbf{23.3} & \textbf{24.3} & \textbf{35.5} \\
    \midrule
    Full & 23.9 & 23.9 & 27.0 & 26.1 & 24.8 & 22.7 & 25.6 & 24.6 & 25.9 & 24.2 & 25.1 & 23.9 & 24.8 & 35.7  \\
    \bottomrule
	\end{tabular*}}
	\subtable{
	\begin{tabular*}{0.99\textwidth}{@{\extracolsep{\fill}}l cc cc || l cc cc cc cc  }
	\toprule
	IWSLT & \multicolumn{4}{c||}{(d) De $\leftrightarrow$ En $\leftrightarrow$ It $\leftrightarrow$ Ro $\leftrightarrow$ Nl}  & MultiUN & \multicolumn{7}{c}{(e) Ar, Ru, Zh $\leftrightarrow$ En} \\
	\midrule
	\multirow{2}{*}{Model} & 
	\multicolumn{2}{c}{De-It} & \multicolumn{2}{c||}{De-Nl} &
	\multirow{2}{*}{Model} &
	\multicolumn{2}{c}{Ar-Ru} & \multicolumn{2}{c}{Ar-Zh} & \multicolumn{2}{c}{Ru-Zh} & Zero & Parallel  \\
	& $\leftarrow$ & $\rightarrow$ & $\leftarrow$ & $\rightarrow$ &
	& $\leftarrow$ & $\rightarrow$ & $\leftarrow$ & $\rightarrow$ & $\leftarrow$ & $\rightarrow$ & Avg & Avg \\
	\midrule 
    PIV-S & 16.7 & 16.3 & -- & --  & PIV-S &\textbf{31.4} & \textbf{33.5} & \textbf{31.2} & \textbf{50.4} & \textbf{31.2} & \textbf{48.0} & \textbf{37.6} & \textbf{48.4}  \\
    PIV-M & \textbf{22.7} & 22.0 & 18.8 & 18.3  & PIV-M & 28.4 & 29.9 & 27.7 & 45.7 & 27.2 & 44.2 & 33.8 & 44.5   \\
    \midrule 
    ZS      & 21.3 & 21.0 & 23.9 & 24.0  & ZS      & 15.6 & 12.7 & 16.7 & 17.0 & 12.8 & 14.9 & 15.0 & 44.5  \\
    ZS+LM & 22.2 & \textbf{22.2} & \textbf{25.0} & \textbf{24.6} &  ZS+LM & 28.0 & 21.5 & 27.3 & 43.8 & 19.9 & 43.3 & 30.6 & 45.8 \\
    ZS+BTTP & -- & -- & -- & -- & ZS+BTTP & \textbf{31.0} & 31.7 & 30.1 & 48.2 & 29.9 & 46.4 & 36.2 & 45.7 \\
    ZS+BTZS& \textbf{22.9} & \textbf{22.9} & \textbf{26.8} & \textbf{26.2} & ZS+BTZS &\textbf{31.4} & \textbf{33.1} & \textbf{31.1} & \textbf{49.4} & \textbf{30.8} & \textbf{46.8} & \textbf{37.1} & \textbf{47.4} \\
    \midrule
    Full & 23.9 & 23.9& 27.0& 26.1& Full & 31.7 & 32.5 & 30.8 & 49.1 & 29.5 & 47.2 & 36.8 & 45.6 \\
    \bottomrule
	\end{tabular*}}

	\caption{Overall BLEU scores including parallel and zero-shot directions on the test sets of three multilingual datasets. In (a) (c) (e), En is used as the pivot-language; no language is available as the pivot for (b); we also present partial results in (d) where a chain of pivot languages are used. For all columns, the highest \textbf{two} scores are marked in bold for all models except for the fully-supervised ``upper bound''. } 
	\label{table:bleu}
\end{table*}

\begin{figure*}[t]
    \centering
    
    \includegraphics[width=\linewidth]{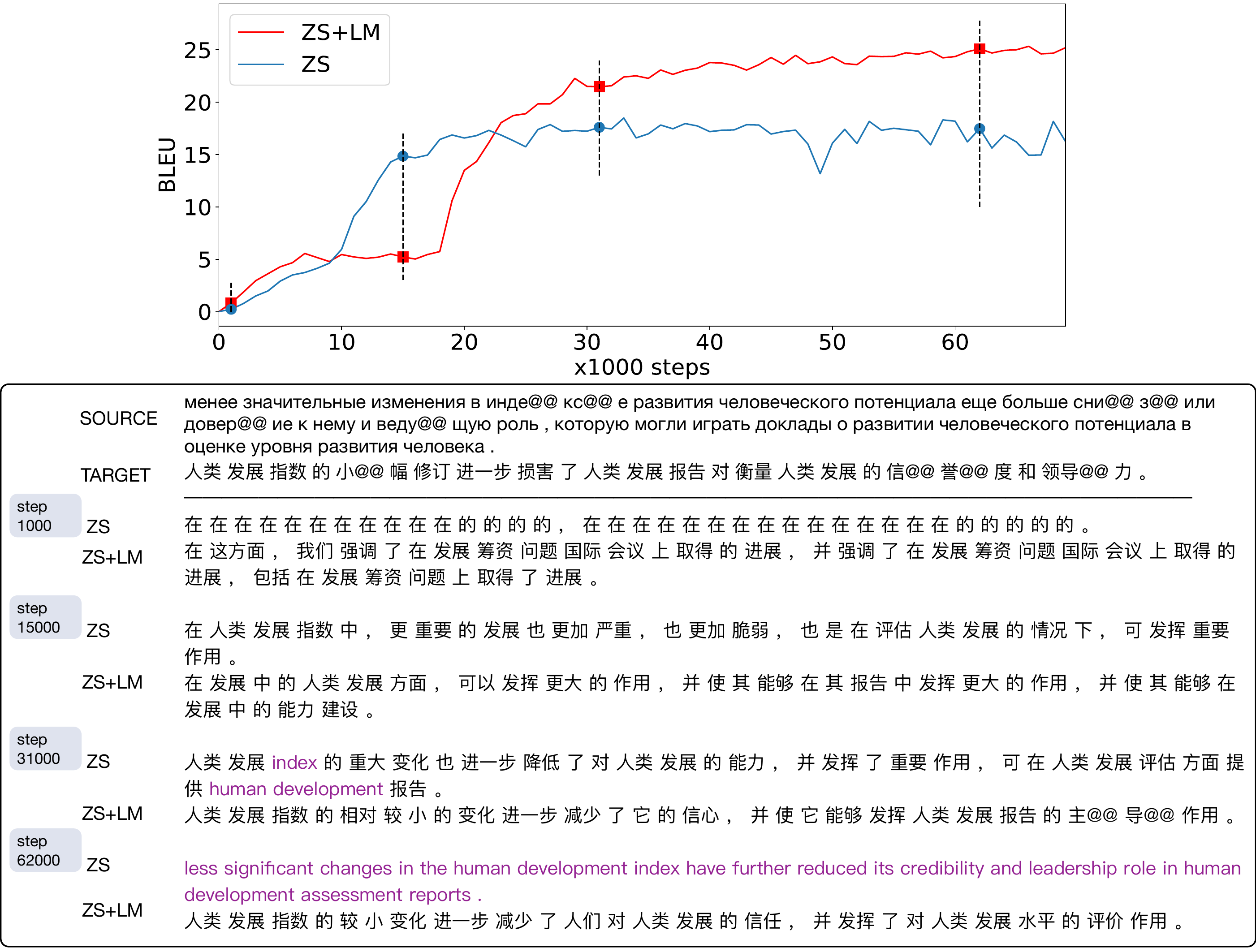}
    \caption{Zero-shot translation performance on Ru $\rightarrow$ Zh from MultiUN dataset. ($\uparrow$) An example randomly selected from the validation set, is translated by both the vanilla zero-shot NMT and that with LM pre-training at four checkpoints. Translation in an incorrect language (English) is marked in pink color. ($\leftarrow$) We showed the two learning curves for the averaged zero-shot BLEU scores on validation set of Multi-UN with the corresponded checkpoints marked.}
    \label{fig.zs_example}
\end{figure*}

\subsection{Results and Analysis}

\paragraph{Overall Performance Comparison}

We show the translation quality of zero-shot NMT
on the three datasets in Table~\ref{table:bleu}. All the results (including pivot-based approaches) are generated using beam-search with beam size $=4$ and length penalty $\alpha=0.6$~\cite{vaswani2017attention}.
Experimental results in Table \ref{table:bleu} demonstrate that both our proposed approaches achieve significant improvement in zero-shot translation for both directions in all the language pairs. 
Only with LM pre-training, the zero-shot NMT has already closed the gap between the performance and that of the strong pivot-based baseline for datasets. 
For pairs which are lexically more similar compared to the pivot language (e.g. Es-Fr v.s. En), ZS+LM achieved much better performance than its pivot-based counterpart.
Depending on which languages we consider, zero-shot NMT with the help of BTTP \& BTZS can achieve a significant improvement around $4\sim 22$ BLEU points compared to the na\"ive approach. For a fair comparison, we also re-implement the alignment method proposed by \newcite{arivazhagan2018missing} based on cosine distance and the results are shown as ZS+Align in Table.~\ref{table:bleu}, which is on average $1.5$ BLEU points lower than our proposed ZS+LM approach indicating that our approaches might fix the degeneracy issue better. 

As a reference of upper bound, we also include the results with a fully supervised setting, where all the language pairs are provided for training. Table \ref{table:bleu} shows that the proposed BTTP \& BTZS are competitive and even very close to this upper bound, and BTZS is often slightly better than BTTP across different languages.

\paragraph{No Pivots}
We conduct experiments on the setting without available pivot languages. Shown in Table \ref{table:bleu}(b), our training sets only contain Es-En and De-Fr. Then if we want to translate from De to Fr, pivot-based methods will not work.
However, we can still perform zero-shot NMT by simply training a multilingual model on the merged dataset. As shown in Table \ref{table:bleu}(a) and (b), although the setting of no pivot pairs performs slightly worse than that with pivot languages, both our approaches (LM, BTZS) substantially improve the vanilla model and achieve competitive performance compared to the fully supervised setting.


\paragraph{A Chain of Pivots}
We analyze the case where two languages are connected by a chain of pivot languages. 
As shown in Table~\ref{table:D1}(IWSLT-b),
we used IWSLT which contains pairs for De-En, En-It, It-Ro, Ro-Nl. 
If we translate from De to Nl with pivot-based translation, pivoting from a chain of languages (De-En-It-Ro-Nl) is required, which suffers from computational inefficiency and error accumulation. 
In such cases, however, zero-shot NMT is able to directly translate between any two languages.
Table \ref{table:bleu}(d) shows that the performance of pivot-based methods dramatically degrades as the length of the chain increases, while ZS does not have this degradation and still achieves large gains compared to the pivot-based translation.

\paragraph{Robustness Analysis}
From Fig.~\ref{fig.robustness}, we show the learning curves of zero-shot NMT with and without our proposed methods. Both the models with LM pre-training and BTZS show robustness in two conditions and achieve competitive and even better results than the pivot-based translation, while the vanilla model is unstable and completely fails after a small number of iterations on \texttt{large-bs}.

\paragraph{Case Study}
We also show a randomly selected example for Ru $\rightarrow$ Zh from the validation set of MultiUN dataset in Fig.~\ref{fig.zs_example}. We can see that at the beginning, the output sentence of ZS+LM is fluent while ZS learns translation faster than ZS+LM. Then, En tokens starts to appear in the output sentence of ZS, and it totally shifts to En eventually.

\section{Related Works}

\paragraph{Zero-shot Neural Machine Translation}
Zero-shot NMT has received increasingly more interest in recent years.
\newcite{platanios2018contextual} introduced the contextual parameter generator, which generated the parameters of the system and performed zero-shot translation.
\newcite{arivazhagan2018missing} conjectured the solution towards the degeneracy in zero-shot NMT was to guide an NMT encoder to learn language agnostic representations.
\newcite{sestorain2018zero} combined dual learning to improve zero-shot NMT.
However, unlike our work, none of these prior works performed quantitative investigation of the underlying cause.

\paragraph{Zero Resource Translation}
This work is also closely related to \textit{zero-resource translation} which is a general task to translate between languages without parallel resources. Possible solutions include \textit{pivot-based} translation, \textit{multilingual} or \textit{unsupervised} NMT.
For instance, there have been attempts to train a single-pair model with a pivot-language~\cite{cheng2016neural,chen2017teacher} or a pivot-image \cite{lee2017emergent,chen2018zero}.

\paragraph{Unsupervised Translation} Unlike the focus of this work, unsupervised translation usually refers to a zero-resource problem where many monolingual corpora are available.
\newcite{lample2017unsupervised,artetxe2017unsupervised} proposed to enforce a shared latent space to improve unsupervised translation quality which was shown not necessary by \newcite{lample2018phrased} in which a more effective initialization method for related languages was proposed.

\paragraph{Neural Machine Translation Pre-training}~ As a standard transfer learning approach, pre-training significantly improves the translation quality of low resource languages by fine-tuning the parameters trained on high-resource languages~\cite{zoph2016transfer,gu2018meta,lample2019cross}.
Our proposed LM pre-training can also be included in the same scope while following a different motivation.

\section{Conclusion}

In this paper, we analyzed the issue of zero-shot translation quantitatively and successfully close the gap of the performance of between zero-shot translation and pivot-based zero-resource translation. We proposed two simple and effective strategies for zero-shot translation. Experiments on the Europarl, IWSLT and MultiUN corpora show that our proposed methods significantly improve the vanilla zero-shot NMT and consistently outperform the pivot-based methods. 
\section*{Acknowledgement}
This research was supported in part by the Facebook Low Resource Neural Machine Translation Award. This work was also partly supported by Samsung Advanced Institute of Technology (Next
Generation Deep Learning: from pattern recognition to AI) and Samsung Electronics (Improving
Deep Learning using Latent Structure). KC thanks
support by eBay, TenCent, NVIDIA and CIFAR.

\bibliography{acl2018}
\bibliographystyle{acl_natbib}

\appendix



\section{Additional Experiments}
\subsection{Trade-off between decoding speed and translation quality}
In Table.~\ref{table:speed}, we empirically tested the decoding speed by using either pivot-based methods or zero-shot NMT. The overhead of switching models in pivot-based translation has been ignored. All the speed are measured as ``ms/sentence'' and tested in parallel on $8$ V100 GPUs using beam-search with a beam size $4$.
\begin{table}[htpb]
	\centering
	\scalebox{1.0}{
	\begin{tabular}{l ccc  }
	\toprule
	Model &  BLEU & Speed \\
	\midrule
    PIV-S \ (greedy)   &  31.1 & 8.3 \\
    PIV-M (greedy)     &  30.6 & 8.3 \\
    PIV-S & 31.5 & 13.3\\
    PIV-M & 31.1 & 13.3\\
    \midrule
    ZS   &  29.4 & \textbf{6.6} \\
    ZS+LM & 32.6 & \textbf{6.6}\\
    ZS+BTTP & 33.1 & \textbf{6.6} \\
    ZS+BTZS & \textbf{33.4} & \textbf{6.6}\\
    \bottomrule
	
	\end{tabular}}
	\caption{Decoding speed and the translation quality (average BLEU scores) of the zero-shot pairs on Europarl dataset.} 
	\label{table:speed}
\end{table}

Vanilla zero-shot NMT is faster but performs worse than pivot-based methods. There exists a trade-off between the decoding speed and the translation quality where we also present a fast pivoting method where we found that using greedy-decoding for the pivot language only affects the translation quality by a small margin. However, both our proposed approaches significantly improve the zero-shot NMT and outperforms the pivot-based translation with shorter decoding time, making such trade-off meaningless.

\subsection{Effect of Using Multi-way Data}
Prior research~\cite{cheng2016neural} also reported that the original Europarl dataset contains a large proportion of multi-way translations. To investigate the affects, we followed the same process in \cite{cheng2016neural,chen2017teacher} to exclude all multi-way translation sentences, which means there are no overlaps in pairwise language pairs. 
The statistics of this modified dataset (Europarl-c) compared to the original Europarl dataset are shown in Table~\ref{table:multi-way dataset}. 
Although we observed a performance drop by using data without multi-way sentences, the results in Table \ref{table:multi-way} show that the proposed LM pre-training is not affected by obtaining multi-way data and consistently improves the vanilla zero-shot NMT. We conjecture that the performance drop is mainly because of the size of the dataset.
Also our methods can easily beat \cite{chen2017teacher} with large margins.

\begin{table}[htpb]
	\centering
	\scalebox{0.87}{
	\begin{tabular}{l | l  r  }
	\toprule
	Dataset & parallel pairs & size/pair  \\
	\midrule
    Europarl   & Es-En, De-En, Fr-En  & 2M  \\
    Europarl-c & Es-En, De-En, Fr-En  & .8M  \\

    \bottomrule
    
	\end{tabular}}
	\caption{Europarl denotes multi-way dataset; Europarl-c denotes non multi-way dataset.} 
	\label{table:multi-way dataset}
\end{table}

\begin{table}[hbp]
    \centering
    \scalebox{0.87}{
	\begin{tabular}{l|cccc }
	\toprule
	\multirow{2}{*}{Model}  & \multicolumn{2}{c}{Es$\rightarrow$Fr} & \multicolumn{2}{c}{De$\rightarrow$Fr} \\
	    &Yes  & No & Yes & No \\
	\midrule 
    PIV-S & 37.95 & 32.98 & 32.20 & 27.94 \\
    PIV-M & 37.15 & 35.08 & 31.46 & 29.78 \\
    \midrule 
    ZS      & 36.69 & 33.22 & 29.59 & 26.91 \\
    ZS + LM & \textbf{40.04} & \textbf{37.22} & \textbf{33.24} & \textbf{30.45} \\
    \midrule
    {\newcite{chen2017teacher}} & $-$ & 33.86 & $-$ & 27.03 \\
    \bottomrule
	\end{tabular}}
	\caption{Effects of multi-way data on Europarl. ``Yes'' means with multi-way translation, and ``No'' means the opposite.} 
	\label{table:multi-way}
\end{table}



\end{document}